\documentclass[11pt,a4paper]{article}
\usepackage[hyperref]{acl2017}

\usepackage{times}
\usepackage{latexsym}
\usepackage[subtle,title=normal,sections=normal,margins=normal,indent=normal]{savetrees}
\usepackage{url}
\usepackage{macros}
\usepackage{algorithmic}

\aclfinalcopy 

\title{Plan, Attend, Generate:\\ Character-Level Neural Machine Translation with Planning}

\author{Caglar Gulcehre\thanks{~~Equal Contribution}\\
  University of Montreal \\
  \And
  Francis Dutil$^{\ast}$ \\
  University of Montreal \\
  \And
  Adam Trischler \\ 
  Microsoft Maluuba \\
  \And
  Yoshua Bengio \\
  University of Montreal}

\date{}

\begin{document}
\maketitle
\begin{abstract}
  We investigate the integration of a planning mechanism into an encoder-decoder architecture with attention for character-level machine translation.
  We develop a model that plans ahead when it computes alignments between the source and target sequences, constructing a matrix of proposed future alignments and a commitment vector that governs whether to follow or recompute the plan. This mechanism is inspired by the strategic attentive reader and writer (STRAW) model. Our proposed model is end-to-end trainable with fully differentiable operations. We show that it outperforms a strong baseline on three character-level translation tasks from WMT'15. Analysis demonstrates that our model computes qualitatively intuitive alignments and achieves superior performance with fewer parameters.
\end{abstract}

\section{Introduction}

Character-level neural machine translation (NMT) is an attractive research problem \citep{lee2016fully,chung2016character,luong2016achieving} because it addresses important issues encountered in word-level NMT. Word-level NMT systems can suffer from problems with rare words\citep{gulcehre2016} or data sparsity, and the existence of compound words without explicit segmentation in certain language pairs can make learning alignments and translations more difficult. Character-level neural machine translation mitigates these issues.

In this work we propose to augment the encoder-decoder model for character-level NMT by integrating a planning mechanism.
Specifically, we develop a model that uses planning to improve the alignment between input and output sequences.
Our model's encoder is a recurrent neural network (RNN) that reads the source (a sequence of byte pairs representing text in some language) and encodes it as a sequence of vector representations; the decoder is a second RNN that generates the target translation character-by-character in the target language. The decoder uses an attention mechanism to align its internal state to vectors in the source encoding.
It creates an explicit plan of source-target alignments to use at future time-steps based on its current observation and a summary of its past actions. At each time-step it may follow or modify this plan.
This enables the model to plan ahead rather than attending to what is relevant primarily at the current generation step. More concretely, we augment the decoder's internal state with (i) an \emph{alignment plan} matrix and (ii) a \emph{commitment plan} vector. The alignment plan matrix is a template of alignments that the model intends to follow at future time-steps,~i.e., a sequence of probability distributions over input tokens. The commitment plan vector governs whether to follow the alignment plan at the current step or to recompute it, and thus models discrete decisions.
This planning mechanism is inspired by the \emph{strategic attentive reader and writer}~(STRAW) of~\citet{vezhnevets2016strategic}.

Our work is motivated by the intuition that, although natural language is \emph{output} step-by-step because of constraints on the output process, it is not necessarily \emph{conceived} and \emph{ordered} according to only local, step-by-step interactions.
Sentences are not conceived one word at a time.
Planning, that is, choosing some goal along with candidate macro-actions to arrive at it, is one way to induce \emph{coherence} in sequential outputs like language.
Learning to generate long coherent sequences, or how to form alignments over long input contexts, is difficult for existing models. NMT performance of encoder-decoder models with attention deteriorates as sequence length increases~ \citep{cho2014properties,sutskever2014sequence}, and
this effect can be more pronounced at the character-level NMT. This is because character sequences are longer than word sequences. A planning mechanism can make the decoder's search for alignments more tractable and more scalable.



We evaluate our proposed model and report results on character-level translation tasks from WMT'15 for English to German, English to Finnish, and English to Czech language pairs. On almost all pairs we observe improvements over a baseline that represents the state of the art in neural character-level translation. In our NMT experiments, our model outperforms the baseline despite using significantly fewer parameters and converges faster in training.

\section{Planning for Character-level Neural Machine Translation}


We now describe how to integrate a planning mechanism into a sequence-to-sequence architecture with attention \citep{bahdanau2014}. Our model first creates a {\it plan}, then computes a soft {\it alignment} based on the plan, and {\it generates} at each time-step in the decoder. We refer to our model as PAG~(Plan-Attend-Generate).

\subsection{Notation and Encoder}

As input our model receives a sequence of tokens, $X = (x_0, \cdots, x_{|X|})$, where $|X|$ denotes the length of $X$. It processes these with the encoder, a bidirectional RNN. At each input position $i$ we obtain annotation vector $\vh_i$ by concatenating the forward and backward encoder states,
$\vh_i = [\vh_{i}^\rightarrow; \vh_{i}^\leftarrow]$, where $\vh_{i}^\rightarrow$ denotes the hidden state of the encoder's forward RNN and $\vh_{i}^\leftarrow$ denotes the hidden state of the encoder's backward RNN.

Through the decoder the model predicts a sequence of output tokens, $Y = (y_1, \cdots, y_{|Y|})$.
We denote by $\vs_t$ the hidden state of the decoder RNN generating the target output token at time-step $t$.

\subsection{Alignment and Decoder}

Our goal is a mechanism that plans which parts of the input sequence to focus on for the next $k$ time-steps of decoding.
For this purpose, our model computes an alignment plan matrix $\mA_t \in \R^{k\times |X|}$ and commitment plan vector $\vc_t \in \R^k$ at each time-step. Matrix $\mA_t$ stores the alignments for the current and the next $k-1$ timesteps; it is conditioned on the current input,~i.e. the token predicted at the previous time-step $\vy_{t}$, and the current context $\psi_t$, which is computed from the input annotations $\vh_i$.
The recurrent decoder function, $f_\text{dec-rnn}(\cdot)$, receives $\vs_{t-1}$, $\vy_{t}$, $\psi_t$ as inputs and computes the hidden state vector
\begin{equation}
\vs_t = f_\text{dec-rnn}(\vs_{t-1}, \vy_{t}, \psi_t).
\end{equation}

Context $\psi_t$ is obtained by a weighted sum of the encoder annotations,
\begin{equation}
	\psi_t = \sum_i^{|X|}\alpha_{ti} \vh_i.
\end{equation}
The alignment vector $\alpha_{t} = \mathtt{softmax}(\mA_{t}[0]) \in \R^{|X|}$ is a function of the first row of the alignment matrix. At each time-step, we compute a candidate alignment-plan matrix $\bar{\mA}_t$ whose entry at the $i^{th}$ row is
\begin{equation}
    \bar{\mA}_t[i] = f_\text{align}(\vs_{t-1},~\vh_j,~\beta^{i}_t,~\vy_{t}),
    \label{eqn:alignment_plan}
\end{equation}
where $f_\text{align}(\cdot)$ is an MLP and $\beta^{i}_t$ denotes a summary of the alignment matrix's $i^{th}$ row at time $t-1$. The summary is computed using an MLP, $f_r(\cdot)$, operating row-wise on $\mA_{t-1}$: $\beta^{i}_t = f_r(\mA_{t-1}[i])$.

The commitment plan vector $\vc_t$ governs whether to follow the existing alignment plan, by shifting it forward from $t-1$, or to recompute it. Thus, $\vc_t$ represents a discrete decision. For the model to operate discretely, we use the recently proposed Gumbel-Softmax trick \citep{jang2016categorical,maddison2016concrete} in conjunction with the straight-through estimator ~\citep{bengio2013} to backpropagate through $\vc_t$.\footnote{We also experimented with training $\vc_t$ using REINFORCE \citep{williams1992} but found that Gumbel-Softmax led to better performance.}
The model further learns the temperature for the Gumbel-Softmax as proposed in \citet{gulcehre2017memory}.
Both the commitment vector and the action plan matrix are initialized with ones; this initialization is not modified through training.

\begin{figure}[ht!]
\centering
\includegraphics[width=0.75\linewidth]{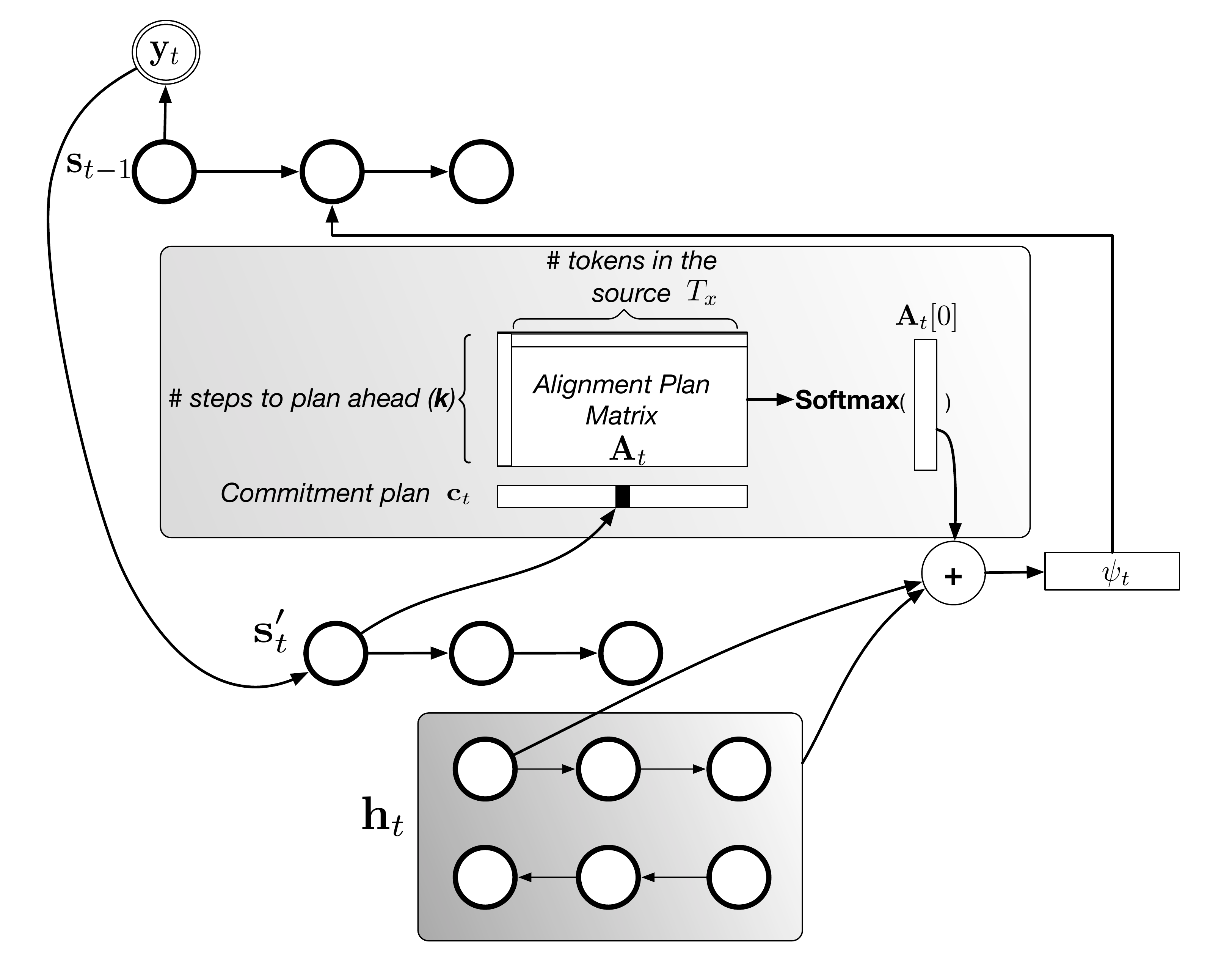}
\caption{Our planning mechanism in a sequence-to-sequence model that learns to plan and execute alignments. Distinct from a standard sequence-to-sequence model with attention, rather than using a simple MLP to predict alignments our model makes a plan of future alignments using its alignment-plan matrix and decides when to follow the plan by learning a separate commitment vector. We illustrate the model for a decoder with two layers $\vs_t^{\prime}$ for the first layer and the $\vs_t$ for the second layer of the decoder. The planning mechanism is conditioned on the first layer of the decoder ($\vs^{\prime}_t$).}
\label{fig:att_planner_nlp}
\end{figure}

\paragraph{Alignment-plan update}

Our decoder updates its alignment plan as governed by the commitment plan. Denoted by $g_t$ the first element of the discretized commitment plan $\bar{\vc}_t$. In more detail, $g_t=\bar{\vc}_t[0]$, where the discretized commitment plan is obtained by setting $\vc_t$'s largest element to 1 and all other elements to 0. Thus, $g_t$ is a binary indicator variable; we refer to it as the commitment switch. When $g_t=0$, the decoder simply advances the time index by shifting the action plan matrix $\mA_{t-1}$ forward via the shift function $\rho(\cdot)$.
When $g_t=1$, the controller reads the action-plan matrix to produce the summary of the plan, $\beta^{i}_t$.
We then compute the updated alignment plan by interpolating the previous alignment plan matrix $\mA_{t-1}$ with the candidate alignment plan matrix $\bar{\mA}_t$. The mixing ratio is determined by a learned update gate $\vu_t \in \R^{k\times|X|}$, whose elements $\vu_{ti}$ correspond to tokens in the input sequence and are computed by an MLP with sigmoid activation, $f_\text{up}(\cdot)$:
\begin{align*}
 	 \vu_{ti} &= f_\text{up}(\vh_i,~\vs_{t-1}), \\ 
     \mA_t[:, i] &= (1 - \vu_{ti}) \odot \mA_{t-1}[:, i] + \vu_{ti} \odot \bar{\mA}_t[:, i].
\end{align*}
To reiterate, the model only updates its alignment plan when the current commitment switch $g_t$ is active. Otherwise it uses the alignments planned and committed at previous time-steps.

\begin{algorithm}
\caption{Pseudocode for updating the alignment plan and commitment vector.}
\begin{algorithmic} \small
    \FOR {$j \in \{1, \cdots |X|\}$}
        \FOR {$t \in \{1, \cdots |Y|\}$}
            \IF {$g_t = 1$}
                \STATE $\vc_t = \text{softmax}(f_{c}(\vs_{t-1}))$
                \STATE $\beta^{j}_t = f_r(\mA_{t-1}[j])$ \COMMENT{Read alignment plan}
                \STATE $\bar{\mA}_t[j] = f_\text{align}(\vs_{t-1},~\vh_j,~\beta^{j}_t,~\vy_{t})$ \COMMENT{Compute candidate alignment plan}
                \STATE $\vu_{tj} = f_\text{up}(\vh_j,~\vs_{t-1},~\psi_{t-1})$ \COMMENT{Compute update gate}
                \STATE $\mA_t~=~(1~-~\vu_{tj}) \odot \mA_{t-1} + \vu_{tj} \odot \bar{\mA}_t$ \COMMENT{Update alignment plan}
            \ELSE
                \STATE $\mA_t = \rho(\mA_{t-1})$ \COMMENT{Shift alignment plan}
                \STATE $\vc_t = \rho(\vc_{t-1})$ \COMMENT{Shift commitment plan}
            \ENDIF
            \STATE Compute the alignment as $\alpha_t = \mathtt{softmax}(\mA_t[0])$
        \ENDFOR
    \ENDFOR
\end{algorithmic}
\label{algo:align_com_plan}
\end{algorithm}


\paragraph{Commitment-plan update}
The commitment plan also updates when $g_t$ becomes 1. If $g_t$ is 0, the shift function $\rho(\cdot)$ shifts the commitment vector forward and appends a $0$-element. If $g_t$ is 1, the model recomputes $\vc_t$ using a single layer MLP ($f_{c}(\cdot)$) followed by a Gumbel-Softmax, and $\bar{\vc}_t$ is recomputed by discretizing $\vc_t$ as a one-hot vector:
\begin{align}
\label{eqn:softmax_out}
\vc_t &= \mathtt{gumbel\_softmax}(f_{c}(\vs_{t-1})), \\
\bar{\vc}_t &= \mathtt{one\_hot}(\vc_t). 
\end{align}

We provide pseudocode for the algorithm to compute the commitment plan vector and the action plan matrix in Algorithm~\ref{algo:align_com_plan}. An overview of the model is depicted in Figure~\ref{fig:att_planner_nlp}.

\subsubsection{Alignment Repeat}

In order to reduce the model's computational cost, we also propose an alternative approach to computing the candidate alignment-plan matrix at every step.
Specifically, we propose a model variant that reuses the alignment from the previous time-step until the commitment switch activates, at which time the model computes a new alignment. We call this variant \textit{repeat, plan, attend, and generate} (rPAG). rPAG can be viewed as learning an explicit segmentation with an implicit planning mechanism in an unsupervised fashion. Repetition can reduce the computational complexity of the alignment mechanism drastically; it also eliminates the need for an explicit alignment-plan matrix, which reduces the model's memory consumption as well. We provide pseudocode for rPAG in Algorithm \ref{algo:align_com_plan}.

\begin{algorithm}
\caption{Pseudocode for updating the repeat alignment and commitment vector.}
\begin{algorithmic} \small
    \FOR {$j \in \{1, \cdots |X|\}$}
        \FOR {$t \in \{1, \cdots |Y|\}$}
            \IF {$g_t = 1$}
                \STATE $\vc_t = \mathtt{softmax}(f_{c}(\vs_{t-1}, \psi_{t-1}))$
                \STATE $\alpha_t = \mathtt{softmax}(f_\text{align}(\vs_{t-1},~\vh_j,~\vy_{t}))$
            \ELSE
                \STATE $\vc_t = \rho(\vc_{t-1})$ \COMMENT{Shift the commitment vector $\vc_{t-1}$}
                \STATE $\alpha_t = \alpha_{t-1}$ \COMMENT{Reuse the old the alignment}
            \ENDIF
        \ENDFOR
    \ENDFOR
\end{algorithmic}
\label{algo:align_com_plan_r}
\end{algorithm}

\subsection{Training}
We use a deep output layer~\cite{pascanu2013construct} to compute the conditional distribution over output tokens,
\begin{equation}
    \label{eq:dec:output}
    p(\vy_t | \vy_{<t}, \vx) \propto \vy_t^{\top}\exp(\mW_o f_o(\vs_t, \vy_{t-1}, \psi_t)),
\end{equation}
where $\mW_o$ is a matrix of learned parameters and we have omitted the bias for brevity. Function $f_o$ is an MLP with $\tanh$ activation.

The full model, including both the encoder and decoder, is jointly trained to minimize the (conditional) negative log-likelihood
\begin{align*}
    \mathcal{L} = - \frac{1}{N} \sum_{n=1}^N \log \; p_{\theta}(\vy^{(n)} | \vx^{(n)}),
\end{align*}
where the training corpus is a set of $(\vx^{(n)}, \vy^{(n)})$ pairs and $\TT$ denotes the set of all tunable parameters. As noted in \citep{vezhnevets2016strategic}, the proposed model can learn to recompute very often which decreases the utility of planning. In order to avoid this behavior, we introduce a loss that penalizes the model for committing too often,
\begin{equation}
    \mathcal{L}_\text{com} = \lambda_\text{com} \sum_{t=1}^{|X|} \sum_{i=0}^{k} ||\frac{1}{k} - \vc_{ti}||_2^2,
\end{equation}
where $\lambda_\text{com}$ is the commitment hyperparameter and $k$ is the timescale over which plans operate.

\begin{figure}[ht!]
\centering 
\includegraphics[width=0.89\linewidth]{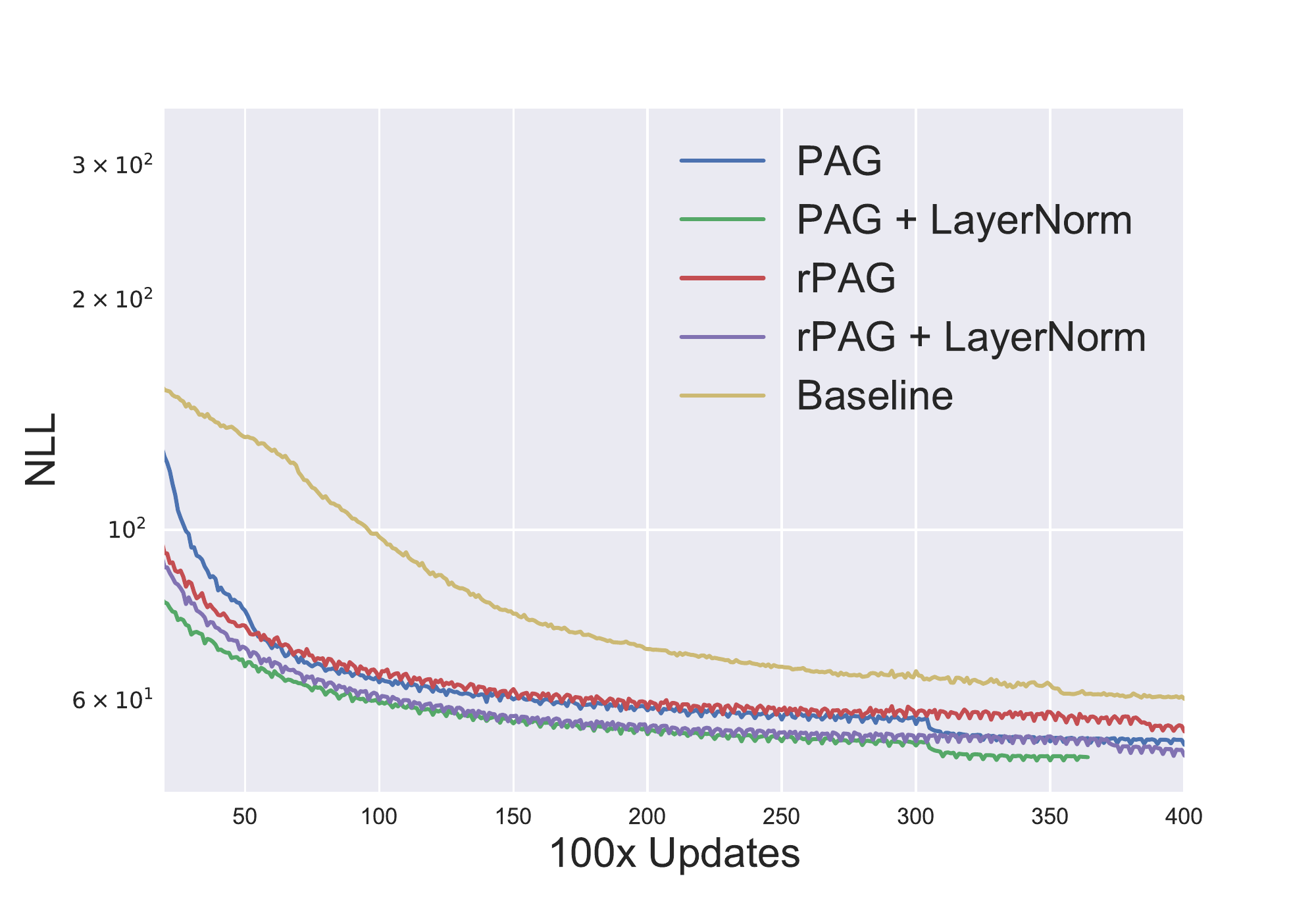}
\caption{Learning curves for different models on WMT'15 for En$\rightarrow$De. Models with the planning mechanism converge faster than our baseline (which has larger capacity).}
\label{fig:learning_curve}
\end{figure}
\vspace{-1mm}

\begin{figure*}[h!]
\begin{subfigure}[b]{0.96\textwidth}
(a)%
\includegraphics[width=\linewidth]{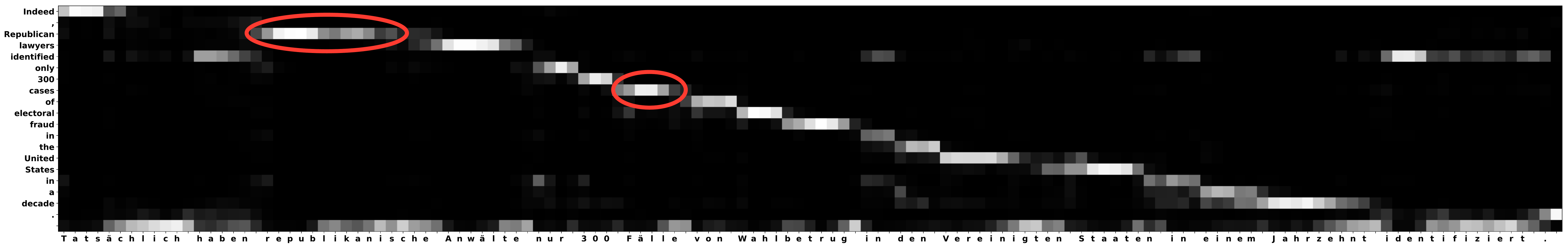}%
\end{subfigure}
\begin{subfigure}[b]{0.96\textwidth}
(b)%
\includegraphics[width=\linewidth]{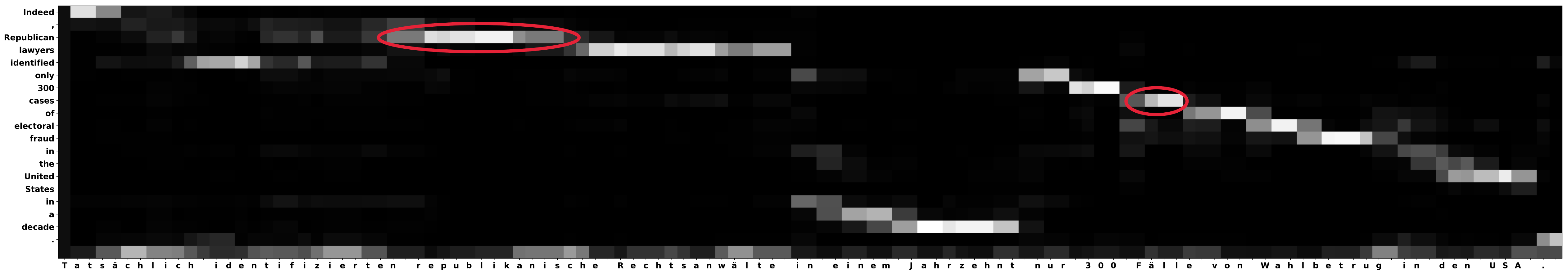}%
\end{subfigure}
\begin{subfigure}[b]{0.96\textwidth}
(c)%
\includegraphics[width=\linewidth]{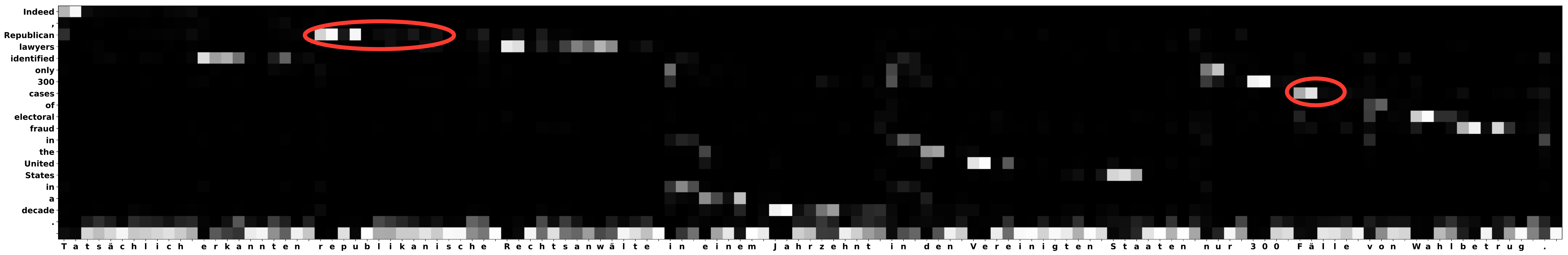}%
\end{subfigure}
\caption{We visualize the alignments learned by PAG in (a), rPAG in (b), and our baseline model with a 2-layer GRU decoder using $\vh_2$ for the attention in (c). As depicted, the alignments learned by PAG and rPAG are smoother than those of the baseline. The baseline tends to put too much attention on the last token of the sequence, defaulting to this empty location in alternation with more relevant locations. Our model, however, places higher weight on the last token usually when no other good alignments exist. We observe that rPAG tends to generate less monotonic alignments in general.}
\label{fig:plan_cmps1}
\end{figure*}

\section{Experiments}


In our NMT experiments we use byte pair encoding (BPE)~\citep{sennrich2015neural} for the source sequence and character representation for the target, the same setup described in~\citet{chung2016character}. We also use the same preprocessing as in that work.\footnote{Our implementation is based on the code available at \url{https://github.com/nyu-dl/dl4mt-cdec}}

We test our planning models against a baseline on the WMT'15 tasks for English to German (En$\rightarrow$De), English to Czech (En$\rightarrow$Cs), and English to Finnish (En$\rightarrow$Fi) language pairs. We present the experimental results in Table \ref{tbl:results_nmt_wmt15}.

As a baseline we use the biscale GRU model of \citet{chung2016character}, with the attention mechanisms in both the baseline and (r)PAG conditioned on both layers of the encoder's biscale GRU ($\vh^1$ and $\vh^2$ -- see \citep{chung2016character} for more detail). Our implementation reproduces the results in the original paper to within a small margin.

Table \ref{tbl:results_nmt_wmt15} shows that our planning mechanism generally improves translation performance over the baseline. It does this with fewer updates and fewer parameters.
We trained (r)PAG for 350K updates on the training set, while the baseline was trained for 680K updates.
We used 600 units in (r)PAG's encoder and decoder, while the baseline used 512 in the encoder and 1024 units in the decoder. In total our model has about 4M fewer parameters than the baseline. We tested all models with a beam size of 15.

As can be seen from Table \ref{tbl:results_nmt_wmt15}, layer normalization \citep{ba2016layer} improves the performance of the PAG model significantly. However, according to our results on En$\rightarrow$De, layer norm affects the performance of rPAG only marginally. Thus, we decided not to train rPAG with layer norm on other language pairs.

\begin{table*}[ht!]
    \small
    \vfill
    \centering
\begin{tabular}{@{}l|l|c||c|c|c@{}}
                       & Model                 & Layer Norm & Dev & Test 2014 & Test 2015 \\ \hline
\multirow{7}{*}{En$\rightarrow$De} & Baseline   &     \xmark     &  21.57   &    21.33       &   {\bf 23.45}        \\
                        &    Baseline$^{\dagger}$   &     \xmark     &  21.4   &   21.16    &    22.1      \\
                       \cline{2-6}
                       & \multirow{2}{*}{PAG}  &    \xmark        & 21.52     &     21.35      &   22.21        \\
                       &                       &  \cmark          & {\bf 22.12}    & {\bf 21.93}          &  22.83         \\ \cline{2-6}
                       & \multirow{2}{*}{rPAG} &    \xmark        &  21.81   &      21.71      &    22.45       \\
                       &                       &   \cmark         &   21.67  & 21.81          &     22.73      \\ \cline{2-6}
                       \hline \hline
\multirow{5}{*}{En$\rightarrow$Cs} & Baseline               &    \xmark        &  17.68   & 19.27          & 16.98           \\  \cline{2-6}
                       & \multirow{2}{*}{PAG}  &    \xmark        &  17.44   &    18.72       & 16.99           \\  
                       &                       &  \cmark          & {\bf 18.78}    &      {\bf 20.9}     &  {\bf 18.59}         \\ \cline{2-6}
                       & rPAG                  &    \xmark        &  17.83   &   19.54        &     17.79      \\ 
                       \hline \hline
\multirow{5}{*}{En$\rightarrow$Fi} & Baseline              &    \xmark        &   11.19  &     -      &  10.93         \\  \cline{2-6}
                       & \multirow{2}{*}{PAG}  &    \xmark        & 11.51    &  -         & 11.13    \\ 
                       &                       &      \cmark      &   {\bf 12.67}    &     -      & {\bf 11.84}  \\ \cline{2-6}
                       & rPAG                  &   \xmark         &  11.50    &  -         &    10.59       \\ \cline{2-6}
                       \hline
\end{tabular}
\vspace{1mm}
\caption{The results of different models on WMT'15 task on English to German, English to Czech and English to Finnish language pairs. We report BLEU scores of each model computed via the \emph{multi-blue.perl} script. The best-score of each model for each language pair appears in bold-face. We use \emph{newstest2013} as our development set, \emph{newstest2014} as our "Test 2014" and \emph{newstest2015} as our "Test 2015" set. $\left(^{\dagger}\right)$ denotes the results of the baseline that we trained using the hyperparameters reported in \cite{chung2016character} and the code provided with that paper. For our baseline, we only report the median result, and do not have multiple runs of our models.}
\label{tbl:results_nmt_wmt15}
\end{table*}


In Figure \ref{fig:plan_cmps1}, we show qualitatively that our model constructs smoother alignments.
In contrast to (r)PAG, we see that the baseline decoder aligns the first few characters of each word that it generates to a byte in the source sequence; for the remaining characters it places the largest alignment weight on the final, empty token of the source sequence. This is because the baseline becomes confident of which word to generate after the first few characters, and generates the remainder of the word mainly by relying on language-model predictions. 
As illustrated by the learning curves in Figure \ref{fig:learning_curve}, we observe further that (r)PAG converges faster with the help of its improved alignments.

\section{Conclusions and Future Work}

In this work, we addressed a fundamental issue in neural generation of long sequences by integrating \emph{planning} into the alignment mechanism of sequence-to-sequence architectures on machine translation problem. We proposed two different planning mechanisms: PAG, which constructs explicit plans in the form of stored matrices, and rPAG, which plans implicitly and is computationally cheaper. The (r)PAG approach empirically improves alignments over long input sequences. In machine translation experiments, models with a planning mechanism outperforms a state-of-the-art baseline on almost all language pairs using fewer parameters. As a future work, we plan to test our planning mechanism at the outputs of the model and other sequence-to-sequence tasks as well.

\clearpage



\bibliography{acl2017}
\bibliographystyle{acl_natbib}

\newpage
\clearpage
\appendix
\onecolumn

\section{Qualitative Translations from both Models}
In Table \ref{tbl:ex_translations}, we present example translations from our model and the baseline along with the ground-truth.
\begin{table*}[h!]
\centering
\caption{Randomly chosen example translations from the development-set.}
\label{tbl:ex_translations}
\small
\begin{tabular}{@{}llll@{}}
\toprule
   & Groundtruth & Our Model (PAG + Biscale)   & Baseline (Biscale) \\ \midrule
1  &
\begin{minipage}[t]{0.33\linewidth}
Eine republikanische Strategie , um der Wiederwahl von Obama entgegenzutreten
\end{minipage} & \begin{minipage}[t]{0.33\linewidth} Eine republikanische Strategie gegen die Wiederwahl von Obama \end{minipage}       & \begin{minipage}[t]{0.33\linewidth}Eine republikanische Strategie zur Bekämpfung der Wahlen von Obama\end{minipage}           \\
2  &\begin{minipage}[t]{0.33\linewidth}  Die Führungskräfte der Republikaner rechtfertigen ihre Politik mit der Notwendigkeit , den Wahlbetrug zu bekämpfen .   \end{minipage}         & \begin{minipage}[t]{0.33\linewidth}  Republikanische Führungspersönlichkeiten haben ihre Politik durch die Notwendigkeit gerechtfertigt , Wahlbetrug zu bekämpfen .   \end{minipage}           & \begin{minipage}[t]{0.33\linewidth} Die politischen Führer der Republikaner haben ihre Politik durch die Notwendigkeit der Bekämpfung des Wahlbetrugs gerechtfertigt . \end{minipage}            \\
3  &  \begin{minipage}[t]{0.33\linewidth}  Der Generalanwalt der USA hat eingegriffen , um die umstrittensten Gesetze auszusetzen .   \end{minipage}             &     \begin{minipage}[t]{0.33\linewidth} Die Generalstaatsanwälte der Vereinigten Staaten intervenieren , um die umstrittensten Gesetze auszusetzen .    \end{minipage}       &    \begin{minipage}[t]{0.33\linewidth}   Der Generalstaatsanwalt der Vereinigten Staaten hat dazu gebracht , die umstrittensten Gesetze auszusetzen .  \end{minipage}         \\
4  &   \begin{minipage}[t]{0.33\linewidth} Sie konnten die Schäden teilweise begrenzen    \end{minipage}            &      \begin{minipage}[t]{0.33\linewidth}   Sie konnten die Schaden teilweise begrenzen  \end{minipage}      &  \begin{minipage}[t]{0.33\linewidth}   Sie konnten den Schaden teilweise begrenzen .  \end{minipage}           \\
5  &   \begin{minipage}[t]{0.33\linewidth}  Darüber hinaus haben Sie das Recht von Einzelpersonen und Gruppen beschränkt , jenen Wählern Hilfestellung zu leisten , die sich registrieren möchten .   \end{minipage}         &  \begin{minipage}[t]{0.33\linewidth} Darüber hinaus begrenzten sie das Recht des Einzelnen und der Gruppen , den Wählern Unterstützung zu leisten , die sich registrieren möchten .    \end{minipage}          &    \begin{minipage}[t]{0.33\linewidth} Darüber hinaus unterstreicht Herr Beaulieu die Bedeutung der Diskussion Ihrer Bedenken und Ihrer Familiengeschichte mit Ihrem Arzt .
    \end{minipage}         \\
\bottomrule
\end{tabular}
\end{table*}\footnote{These examples are randomly chosen from the first 100 examples of the development set. None of the authors of this paper can speak or understand German.}

\end{document}